\documentclass[10pt,twocolumn,letterpaper]{article}

\usepackage{iccv}
\usepackage{times}
\usepackage{epsfig}
\usepackage{graphicx}
\usepackage{amsmath}
\usepackage{amssymb}
\usepackage{subfigure}
\usepackage{color}
\usepackage{booktabs}
\usepackage{multirow}
\usepackage{bm}
\usepackage{tablists}

\usepackage[breaklinks=true,bookmarks=false]{hyperref}

\iccvfinalcopy 

\def\iccvPaperID{****} 

\ificcvfinal\fi

\begin{document}

\title{Semi-supervised Contrastive Learning with Similarity Co-calibration}

\author{Yuhang Zhang\\
Shanghai Jiao Tong University\\
{\tt\small hang\_universe@sjtu.edu.cn}
\and
Xiaopeng Zhang\\
Huawei Inc.\\
{\tt\small zxphistory@gmail.com}
\and
Robert.C.Qiu\\
Shanghai Jiao Tong University\\
{\tt\small rcqiu@sjtu.edu.cn}
\and
Jie Li\\
Shanghai Jiao Tong University\\
{\tt\small lijiecs@sjtu.edu.cn}
\and
Haohang Xu\\
Shanghai Jiao Tong University\\
{\tt\small xuhaohang@sjtu.edu.cn}
\and
Qi Tian\\
Huawei Inc.\\
{\tt\small tian.qi1@huawei.com}
}

\maketitle
\ificcvfinal\thispagestyle{empty}\fi

\begin{abstract}
     Semi-supervised learning acts as an effective way to leverage massive unlabeled data. In this paper, we propose a novel training strategy, termed as \textbf{Semi-supervised Contrastive Learning (SsCL)}, which combines the well-known contrastive loss in self-supervised learning with the cross entropy loss in semi-supervised learning, and jointly optimizes the two objectives in an end-to-end way. The highlight is that different from self-training based semi-supervised learning that conducts prediction and retraining over the same model weights, SsCL interchanges the predictions over the unlabeled data between the two branches, and thus formulates a co-calibration procedure, which we find is beneficial for better prediction and avoid being trapped in local minimum. Towards this goal, the contrastive loss branch models pairwise similarities among samples, using the nearest neighborhood generated from the cross entropy branch, and in turn calibrates the prediction distribution of the cross entropy branch with the contrastive similarity. We show that SsCL produces more discriminative representation and is beneficial to few shot learning. Notably, on ImageNet with ResNet50 as the backbone, SsCL achieves $\bm{60.2\%}$ and $\bm{72.1\%}$ top-1 accuracy with $1\%$ and $10\%$ labeled samples, respectively, which significantly outperforms the baseline, and is better than previous semi-supervised and self-supervised methods.
\end{abstract}

\section{Introduction}
\begin{figure}
    \centering
    \includegraphics[width=\columnwidth]{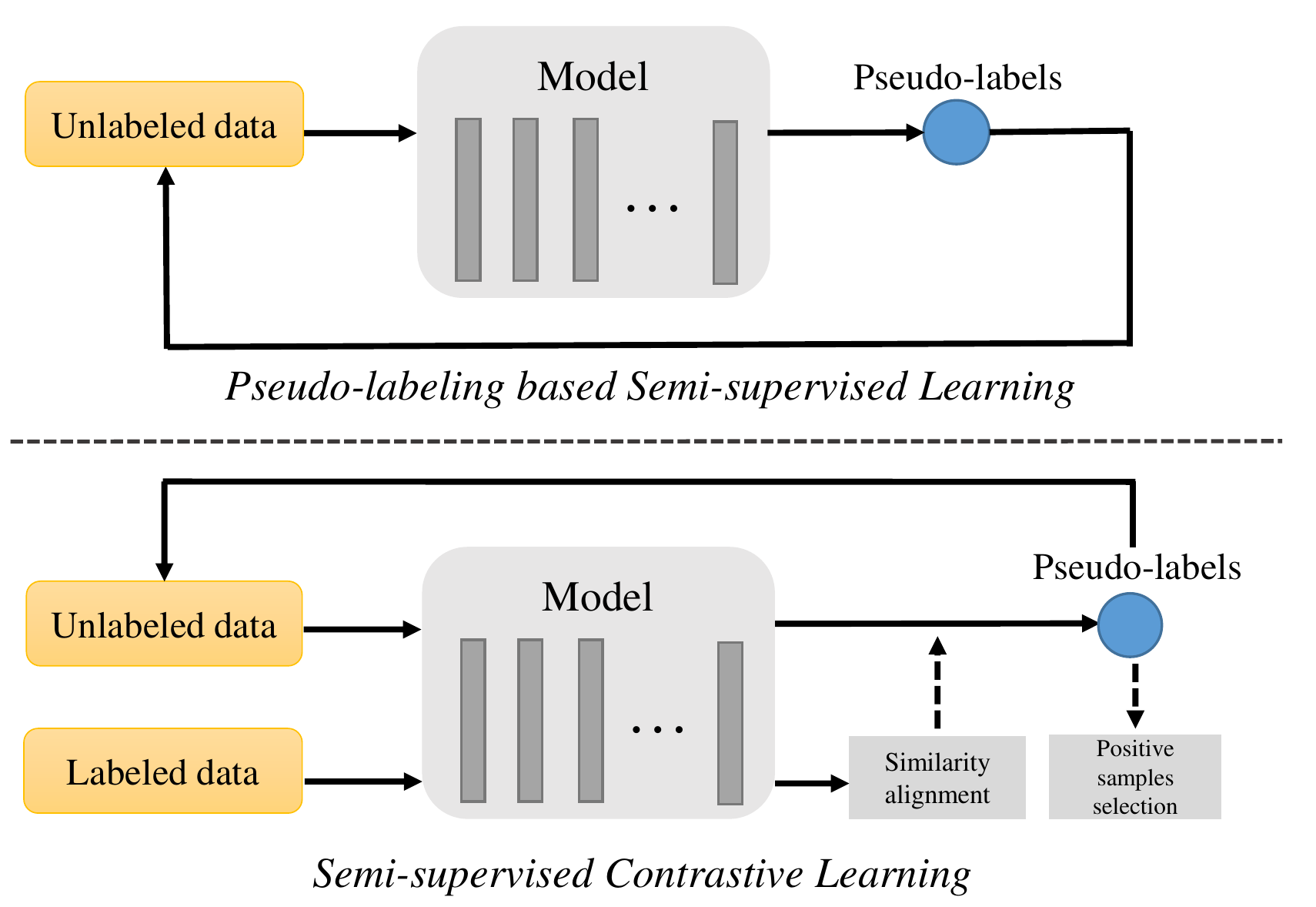}
    \caption{Comparisons between cross-entropy based semi-supervised learning and our proposed SsCL. The highlight is that we co-calibrate the predictions over the two branches and jointly train the two objectives in an end-to-end way.}
    
    \vspace{-0.5cm}
    \label{fig:advl}
\end{figure}

Semi-supervised learning(SSL) has attracted more attention over the past years, as it allows for training network with limited labeled data to improve the model performance with the availability of large scale unlabeled data\cite{arazo2020pseudo,berthelot2019remixmatch,lee2013pseudo,2019Virtual,rasmus2015semi,zhai2019s4l}. A general pipeline of SSL is based on pseudo-labeling and consistency regularization\cite{arazo2020pseudo,lee2013pseudo,sohn2020fixmatch}. The key idea is to train the model with limited labeled data and predict the pseudo labels for the unlabeled samples. In this pipeline, the images with high-confidence prediction are selected for retraining, using either one-hot hard labels or the predicted distribution of soft labels. In order to avoid homogenization in SSL, \emph{i.e.,} predicting and retraining with the same samples, data augmentation is widely used, and has been demonstrated to be particularly effective when combined with consistency regularization, which penalizes the discrepancy between different augmentations of the same samples. In SSL, the network is usually optimized by minimizing the cross entropy loss between the prediction and the target distribution as fully supervised learning. However, as pointed out by \cite{sukhbaatar2014training,zhang2018generalized}, such cross entropy loss is not robust to noisy labels, which is inevitable in semi-supervised learning. This is especially true for large scale dataset such as ImageNet, where limited labeled samples cannot represent the specific class properly, and cross entropy based semi-supervised methods \cite{berthelot2019remixmatch,2019Virtual,sohn2020fixmatch,zhai2019s4l} suffer limited performance gain.

On the other hand, recent advances in self-supervised learning have demonstrated powerful performance in semi-supervised learning scenarios, especially with the achievements of contrastive learning \cite{he2020momentum}. In this setting, the model is first pretrained with large scale unlabeled data, and then fine-tuned with few shot labeled samples. In the pretraining stage, a typical solution is to treat each image as well its augmentations as a separate class, and the features among different transformations of an image are pulled closer together, while all other instances are treated as negatives and pushed away. However, due to the lack of labels, the self-supervised pretraining stage is task agnostic, which is not optimal for specific tasks. 



In this paper, we propose a novel training strategy that incorporates contrastive loss into semi-supervised learning, and jointly optimize the two objectives in an end-to-end way. The motivation is that it should be better for specific task when we incorporate class specific priors into the contrastive pretraining stage, and hope that the two different feature learning strategies would benefit each other for better representation. Towards this goal, we introduce a co-calibration mechanism to interchange predictions between the two branches, one branch based on cross entropy loss and the other based on contrastive loss. In particular, the pseudo labels generated from the cross entropy loss term are used to search for nearest neighbors for contrastive learning, while the similarity embedding learned by the contrastive term is used in turn to adjust the prediction in the pseudo labeling branch. In this way, the information between the two branches is interflowed, which is complementary with each other and thus avoid being trapped in the local minimum course by self-training. 


In order to facilitate co-calibration between the two branches, we extend the contrastive loss \cite{he2020momentum} to allow for multiple positive samples during each forward propagation, and pulling samples that are similar to the specified class prototype for discriminative representation. In order to tackle the noisy samples especially at the initial training stage, we propose a self-paced learning strategy that adaptively adjusts the loss that the mined samples belonging to the assigned class. In this way, the model is gradually evolving as the similarity computation becomes more accurate. Furthermore, in order to avoid the biased representation of each class with limited labeled data, we introduce a data mix strategy to enrich samples of each class. Specifically, we periodically select similar samples during the semi-supervised learning process, and apply data mixing \cite{zhang2018mixup} operation over the randomly selected two samples from the ground truth and similar sample pools, respectively. In this way, we are able to enrich the class specific samples in a robust and smooth way. 
 
 Integrating contrastive loss into semi-supervised learning produces a much more powerful framework, and experiments demonstrate that the enriched representation is beneficial for data mining and results in improved representation. Notably, using a standard ResNet-50 as the backbone, we achieve $60.2\%$ and $72.1\%$ top-1 accuracy on the large scale ImageNet dataset with $1\%$ and $10\%$ labeled samples, respectively, which consistently surpasses the semi-supervised and self-supervised methods.

\section{Related Work}
Our approach is most related with recent advances covering semi-supervised learning and self-supervised learning. We briefly review related works and clarify the differences btween them and our method.
\paragraph{Semi-supervised Learning} makes uses of few shot labeled data in conjunction with a large number of unlabeled data\cite{berthelot2019remixmatch,guillaumin2010multimodal,long2020self,kingma2014semi,oliver2018realistic,sohn2020fixmatch,rasmus2015semi,tarvainen2017mean}.
Pseudo labeling is a widely used method in early works, which uses the predictions of the model to label images and retain the model with these predictions \cite{lee2013pseudo,arazo2020pseudo}. Then \cite{sohn2020fixmatch} combines consistency regularization and pseudo-labeling to align the predictions between weakly and strongly augmented unlabeled images. S$^{4}$L\cite{zhai2019s4l} is the first unified work that combines self-supervised learning with semi-supervised learning. As to our approach, we not only use pseudo labeling method to build a part of our loss function,  but also lets the pseudo label play an important role in our multi-positive contrastive learning, and contrastive learning representations will also help the model obtain more accurate pseudo labels. 

\paragraph{Contrastive Learning} in recent works mainly benefits from instance discrimination\cite{wu2018unsupervised}, which regards each image and its augmentations as one separate class and others are negatives \cite{he2020momentum,chen2020simple,dosovitskiy2015discriminative,chen2020improved, hjelm2018learning, oord2018representation, tian2019contrastive, grill2020bootstrap}. 
\cite{wu2018unsupervised} use a memory bank to store the pre-computed representations from which positive examples are retrieved given a query. Based on it, \cite{he2020momentum} used a momentum update mechanism to maintain a long queue of negative examples for contrastive learning, while \cite{chen2020simple} used a large batch to produce enough negative samples. 
These works prove that contrastive learning reaches better performance on learning characteristics of data. Previous contrastive based semi-supervised learning works are almost two-stage ones, \emph{i.e.,} using contrastive learning to pretrain a backbone and then using few shot labeled data to fine-tune it. In contrast, our method trains the model in an end-to-end way, which is able to better make use of the advantages of the features learned by different loss functions and class specific priors.

\begin{figure*}
	\centering
	\includegraphics[width=\textwidth,height=7cm]{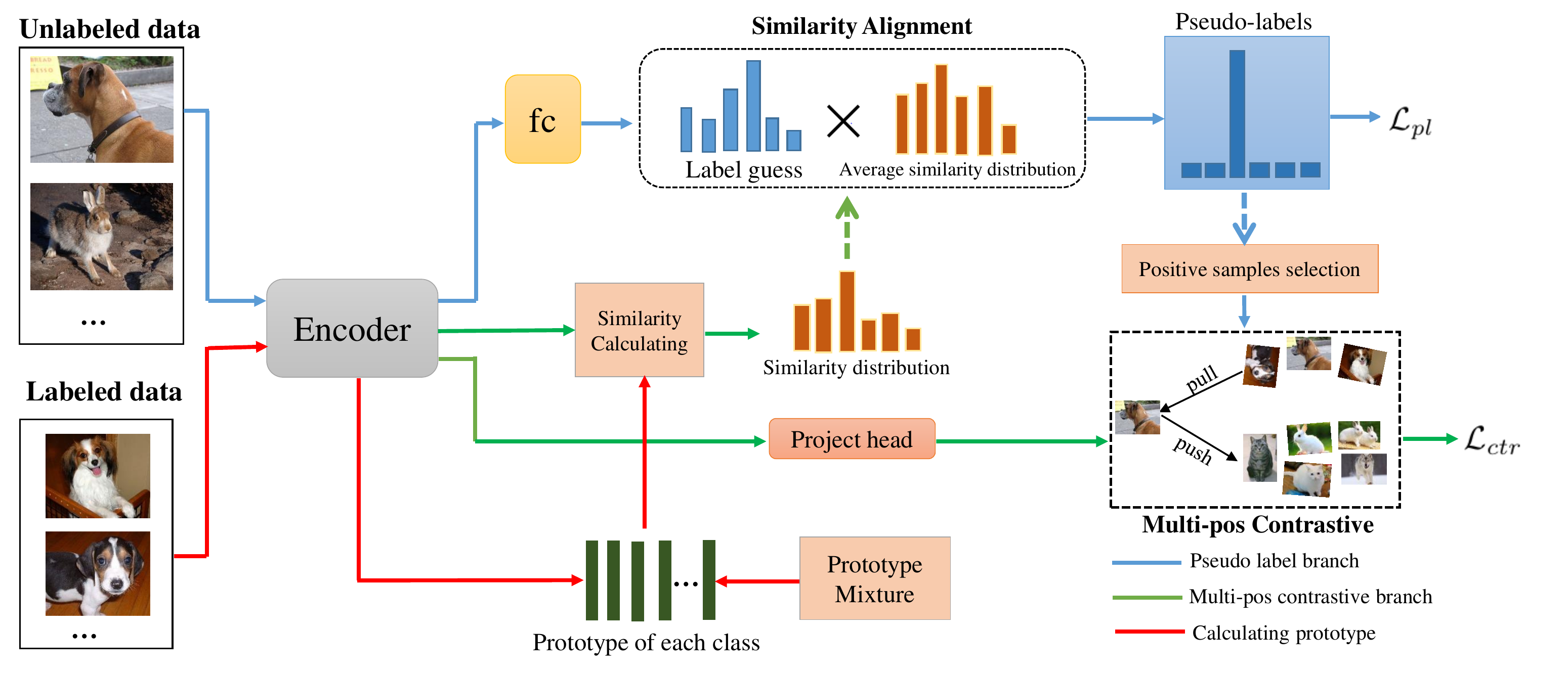}
	\caption{An overview of our proposed semi-supervised contrastive learning framework. Different from traditional contrastive learning in unsupervised paradigm, the few shot labeled data is utilized for contrastive learning, which targets at learning more discriminative representation in an end-to-end way.}
	\vspace{-0.4cm}
	\label{fig:pipeline}
\end{figure*}

\section{Method}
We first give an overview of the proposed Semi-supervised Contrastive Learning framework. SsCL includes a combination of two well-known approaches, \emph{i.e,} the pseudo labeling strategy with cross entropy loss, and the instance discrimination with contrastive loss. The highlight is that we jointly optimize the two losses with a shared backbone in an end-to-end way, and importantly, the pseudo labels between the two branches are calibrated in a co-training mechanism, which we find is beneficial for better prediction over the unlabeled data. 

Specifically, denote $\mathcal{X} = \{ (x_b, y_b): b\in (1,...,B) \}$ as a batch of \emph{B} labeled samples, where $y_b$ denotes labels, and $\mathcal{U} = \{ u_b: b\in (1,...,\mu B) \}$ as a batch of $\mu\emph{B}$ unlabeled samples, where $\mu$ determines the relative size of $\mathcal{X}$ and $\mathcal{U}$. $DA$ represents a random data augmentation conducted over an image. Overall, SsCL targets at optimizing three losses: 1) the supervised loss $\mathcal{L}_{sup}$ optimized over the labeled data; 2) the pseudo labeling loss $\mathcal{L}_{pl}$ penalized on the unlabeled data and 3) the contrastive loss $\mathcal{L}_{ctr}$ that enforces pairwise similarity among neighborhood samples. The whole framework is shown in Fig. ~\ref{fig:pipeline}, and we omit the supervised loss term for brevity. 

The supervised loss is simply conducted over the labeled data $x_b$ with cross-entropy minimization, using the ground truth labels $y_b$:
\begin{equation}\label{loss_sup}
    \mathcal{L}_{sup} = \frac{1}{B}\sum_{b=1}^B H(y_b,p(y|DA(x_b)))
\end{equation}

Similarly, the pseudo labeling loss is penalized over the unlabeled data $u_b$, using the pseudo labels $\hat{p}_b$: 
\begin{equation}\label{loss_pl}
	\mathcal{L}_{pl} = \frac{1}{\mu B}\sum_{b=1}^{\mu B}\textbf{1} (max~\hat{p_b}\ge \tau )H(\hat{p_b},p(y|DA(u_b))),
\end{equation}
where $\hat{p}_b$ denotes the model's prediction on the unlabeled samples, which is obtained by calibrating the pseudo labels $p_b$ predicted from the cross entropy loss term with similarity distribution $\hat{p}_s$ from the contrasstive loss term, and will be elaborated in the following section. $\bm{1}$ is an indicator function and we only retain $p_b$ whose largest class probability is above a certain threshold $\tau$ for optimization.



For contrastive loss, different from \cite{he2020momentum}, we adjust the contrastive loss term to allow for multiple positive samples during each forward propagation so that similar images that belong to the corresponding class prototype are pulled together for more discriminative representation. We defer the definition of contrastive loss term in the following section. The overall losses can be formulated as: 

\begin{equation}
    \mathcal{L} = \mathcal{L}_{sup} + \lambda_{pl} \mathcal{L}_{pl} + \lambda_{ctr} \mathcal{L}_{ctr},
\label{E:loss_total}
\end{equation}
where $\lambda_{pl}$ and $\lambda_{ctr}$ are the balancing factors that control the weights of the two losses. In the  following, we describe the terms $\mathcal{L}_{pl}$ and $\mathcal{L}_{ctr}$ in detail and elaborate the similarity co-calibration procedure among the two branches. 

\subsection{Pseudo Label Calibration with Contrastive Similarity}

For conventional semi-supervised learning with cross entropy loss, the pseudo label $p_b$ of an unlabeled sample $u_b$ is simply derived by predicting distribution on $DA_a(u_b)$ using the current model, and enforces the cross-entropy loss against the model’s output for $DA_b(u_b)$ in the following training procedure. However, the pseudo labeling and re-training is conducted over the same network, which suffers from model homogenization issue and is easy to be trapped in local minimum. In this section, we propose a pseudo label calibration strategy that refines the prediction $p_b$ via the similarity distribution $p_s$ from the contrastive loss term. 
\paragraph{Similarity Distribution} The similarity distribution $p_s$ over the classes is obtained as follows. Given a list of $n$ few shot labeled samples $\bm{X}_c=\{x_i\}_{i=1}^n$ for class $c$, we compute the feature representation $\bm{v}_i$ of $x_i$ with current encoder $E_q$ 
where $\bm{v}_i=E_q(x_i)$. The feature representation of class $c$ is simply obtained by averaging the features, \emph{i.e.,} $\bm{\tilde{v}}_c=\sum \bm{v}_i/nZ$, where $Z$ is a normalizing constant value. Given an unlabeled sample $x'$ with normalized feature $\bm{\tilde{v'}}$, we can get its cosine similarity distribution $p_s=\{s_1,s_2...,s_c  \}$ with $C$ predefined prototypes, where $s_k$ denotes the similarity between ($\bm{\tilde{v'}}$, $\bm{\tilde{v}}_k$). 

\begin{figure} 
	\centering
	\includegraphics[width=\columnwidth,height=4.5cm]{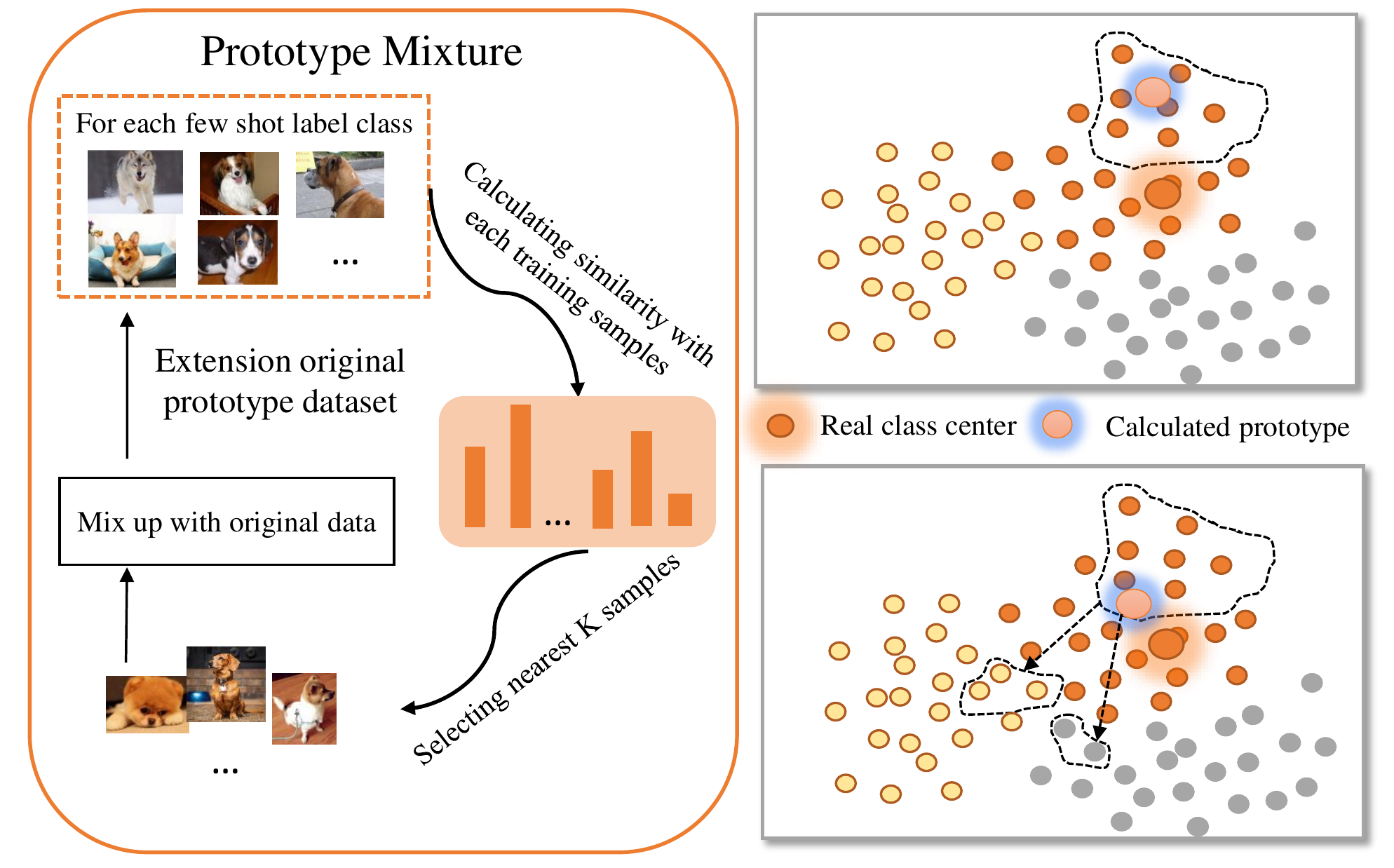}
	\caption{An illustration of the proposed prototype mixture module. Class specific prototypes are enriched by mixing samples that from the mined similar images and those with ground truth labels.}
	\vspace{-0.4cm}
	\label{fig:mixture}
	
\end{figure}

Inspired by \cite{berthelot2019remixmatch}, we introduce a form of fairness prediction $\hat{p}_s$ by maintaining a running average of distribution $p_s$ (average vector of $p_s$ over the last $t$ batches, here we simply set $t=128$) over the course of training. Overall, $\hat{p}_s$ can be treated as a global adjustment among the scores for each class, and thus avoid biased prediction $p_b$ that conducting over a single image $u_b$. Then the calibration is simply conducted by scaling $p_b$ with $\hat{p_s}$ and then re-normalizing to form a valid probability distribution $\hat{p}_b$. We use $\hat{p}_b$ as calibrated pseudo labels for optimization following Eq. \ref{loss_pl}.
\paragraph{Prototype Refinement} The similarity distribution $p_s$ requires a predefined class specific prototype generated by the few labeled samples, and each prototype $\bm{\tilde{v}}_c$ is simply computed using the available ground truth labels belonging to class $c$. However, this representation is easily biased to the few shot samples, which is insufficient to model the real distribution. 
To solve this issue, we propose a prototype mixture strategy to refine the feature representation of each class in a robust way. The new samples are generated via a simple mixup \cite{zhang2018mixup} operation, conducted between the mined samples and those with ground truth labels. In this way, based on the continuity of feature space, we are able to ensure that the generated samples do not drift away too much from the real distribution of that class, while enriching the feature representation in a robust way. Specifically, given the few shot labeled sample set $\bm{X}_c$ for class $c$ and a list of nearest neighborhood unlabeled samples $\Omega_c$ that belong to class $c$, we randomly select a ground truth sample $x_c \in \bm{X}_c$, and a nearest sample $x_{nearest} \in $ $\Omega_c$, and mix them to generate a new sample $\hat{x}$:
\begin{equation}
    \hat{x} = \lambda\cdot x_{c} + (1-\lambda)\cdot x_{nearest},
\end{equation}
where $\lambda$ is the combination factor and is sampled from beta distribution Beta$(\alpha,\alpha)$ with parameter $\alpha$. In our implementation, we simply set $\alpha=1$, and $\hat{x}$ will be added to $\bm{X}_c$ only use for calculating prototype for class $c$.
The purpose of this step is to smooth the noise caused during the sample selection process. We visualize the benefits of this step in Fig.~\ref{fig:mixture}. Benefiting from the mixed samples, the feature representation is refined to better represent the class prototype. In the experimental section, we would validate its effectiveness.

\subsection{Contrastive Learning with Pseudo Label Propagation}
In contrastive learning, the positive samples are simply constrained within a single image with different data transformations, and all other images are treated as negatives and pushed away. In this section, we incorporate the few shot labeled priors into contrastive learning, and target at pursuing more discriminative representation with the help of the mined samples from the pseudo labeling branch. Specifically, we extend the original contrastive learning loss in \cite{he2020momentum} and pull the mined samples closer together with the predefined prototypes. Benefiting from the class specific priors, the feature representation is more discriminative, which is beneficial for semi-supervised learning that encourages class separability and better representation.

\paragraph{Loss Function}
In order to efficiently make use of the unlabeled samples in a robust way, inspired by \cite{sun2020circle}, we adjust the original contrastive loss from \cite{he2020momentum} in three aspects. First, we extend it to allow for multiple positives for each forward propagation; second, we introduce a margin value $m$ which enforces class separability for better generalization, which we find is robust for data mining; third, considering that the mined samples may not be reliable, especially during the initial training stages, we introduce a soften factor $\alpha_p$ for each mined positive sample, and adaptively penalize the discrepancy according to its similarity to the corresponding class. For ease of expression, we make a deformation of original contrastive loss by using $s_p$ to represent the similarity of positive samples while $s_n$ represents the similarity of negative samples, and denote $\gamma=1/\tau$. After introducing margin $m$ and soften factor $\alpha_p$, the loss $\mathcal{L}_{ctr}$ can be reformulated as:

\begin{footnotesize}
\begin{align}\label{circle_contrastive}
\mathcal{L}_{ctr} &=  - \log \left(\frac{{\sum\limits_{{k^ + }}\exp (\gamma(\alpha_p{s_p} - m))}}{{\sum\limits_{{k^ + }}\exp (\gamma(\alpha_p {s_p} - m)) + \sum\limits_{{k^ - }} {\exp (\gamma{s_n})} }}\right) \notag \\
&=  - \log \left(\frac{{\sum\limits_ {{k^ + }}\exp (\gamma\alpha_p {s_p})}}{{\sum\limits_{{k^ + }}\exp (\gamma\alpha_p {s_p}) + \sum\limits_{{k^ - }} {\exp (\gamma({s_n} + m))} }}\right) \notag \\
&= \log \left(1 + \sum\limits_{{k^ - }} {\exp (\gamma({s_n} + m))} \sum\limits_{{k^ + }}\exp ( - \gamma \alpha_p {s_p})\right).
\end{align}
\end{footnotesize}

Designed with the above three properties, the loss is well suited for semi-supervised learning. In the following, we would analyze its advantages in detail.
\paragraph{Positive Sample Selection} For this branch, the positive samples are simply selected from the pseudo label loss branch. 
Specifically, given an unlabeled sample $u_b$ with pseudo labels $p_b$, which is predicted by the $fc$ layers of the cross entropy term. We assign it to class $c$ if $argmax \, \, p_b = c$, denoting that this sample is most similar with class $c$. Then we pulled $u_b$ together with those samples that have ground truth label $c$. In this way, the unlabeled samples are pulled towards the specific instances corresponding to that class, and the representation is more compact. In practice, all the positive samples are maintained in the key encoder $E_k$. Benefit from the moment contrast mechanism in MoCo\cite{he2020momentum}, the encoder $E_k$ is asynchronously updated and add more positive keys doesn't increase the computation complexity.

\paragraph{Self-paced Weighting Mechanism}
A critical issue when using cosine similarity for positive sample selection is that at the initial training stages, the model is under-fitted, and the accuracy of using the nearest neighbor is very low. As a result, the incorrect positive samples would be assigned to a class, causing the model to update in the wrong direction, which is harmful for generalization. To solve this issue, we propose a self-paced learning method to enhance the optimization flexibility and it is robust to the noisy samples. This is achieved by assigning each selected positive sample a weight factor $\alpha_p$, which is determined by the similarity between the current sample and the predefined most similar class prototype. It is equivalent to adding a penalty term to the loss function, and with the increased accuracy of positive sample selection, the penalty becomes larger, and in turn enforces the compact representation.
\paragraph{Analysis of Our Loss Function}We now step deep into the optimization objective Eq. (\ref{circle_contrastive}) to better understand its properties. Towards this goal, we first make some approximate transformations using the following formulations:




\begin{align}
\label{eq:3} \mathcal{L}_{LSE}(x;\gamma) &= \frac{1}{\gamma}\log \sum\limits_{i}exp(\gamma x_i)\approx max(x) \\
\label{eq:4} \mathcal{L}_{-LSE}(x;\gamma) &=-\frac{1}{\gamma}\log \sum\limits_{i}exp(-\gamma x_i)\approx min(x) \\
\label{eq:5} \mathcal{L}_{Softplus}&=\log(1+exp(x))\approx max(x,0)
\end{align}

From above formulas, we can transform the Eq.~ (\ref{circle_contrastive}) to:
\begin{align}
    \mathcal{L}_{ctr}  \approx max(\gamma(max(s_n)-min(\alpha_p s_p)+m),0)
    \label{eq:min_max}
\end{align}

The specific derivation process will be shown in the Supplementary. From the above equations, we have the following observations: (1) According to Eq.~(\ref{eq:min_max}), the objective is to optimize the upper bound of positive pairs and lower bound of negative pairs, and introducing of $\alpha_p$ makes this process more flexible; (2) Based on the properties of Eq.~(\ref{eq:3}) and Eq.~(\ref{eq:4}), whose gradient is the $softmax$ function, the positive and negative pairs obtain equal gradient, regardless of the number of positive and negative sample pairs. This gradient equilibrium ensures the success of our multi-positive sample training.

\section{Experiment}
In this section, we evaluate our method on several standard image classification benchmarks, including CIFAR-10\cite{AlexLearning2009} and ImageNet\cite{2009ImageNet} when few shot labels are available. We also evaluate our representations by transferring to downstream tasks.

For the contrastive branch, we make use of two encoders follow \cite{he2020momentum}, one for training(named as $E_q$) and the other one for momentum update (named as $E_k$) to produce negative keys. For prediction co-calibration, the contrastive similarity is computed via features obtained by $E_q$ and the prediction on unlabeled data is achieved by the $fc$ head. For efficiency, we conduct these operations offline, and update the pseudo labels (for selecting positive samples) and prototype for each category every 5 epochs during the training process. We find that the results are robust as the update frequency ranges from 1 to 20. Besides, in practice, the similarity computation and pseudo label inference can be done within only a few minutes, and brings about negligible costs comparing with the contrastive learning iteration. 

As for prototype refinement, we periodically update the feature representation of each class using the mixed samples for every $5$ epochs, the same frequency as the similarity computation, and add fixed number of samples to each class. The number of samples is simply set equal to the labeled number of each class, hoping that the mixed samples do not overwhelm the ground truth samples, and we find that the result is robust for a range of mixed samples when they are at the same scale with the ground truth labels.


\subsection{CIFAR-10}
We first compare our approach with various methods  on CIFAR-10, which contains 50000 images with size of 32 $\times$ 32. We conduct the experiments with different number of available labels, and each setting is evaluated on 5 folds. 
\paragraph{Implementation Details} As recommended by \cite{sohn2020fixmatch}, we use Wide ResNet-28-2\cite{zagoruyko2016wide} as backbone, as well as training protocol, $\mu=4$, $B=64$, $\tau = 0.95$, and all loss weights (e.g. $\lambda_{pl}$) are set as 1. The projection head is a 2-layer MLP that outputs 64-dimensional embedding. The models are trained using SGD with a momentum of 0.9 and a weight decay of 0.0005. We train our model for 200 epochs, using learning rate of 0.03 with a cosine decay schedule. For the additional hyperparamaters, we set the number of negative samples $K=4096$, $\gamma=5$.
Besides, we use two different augmentations on unlabeled data, The $DA_a$ here is the standard crop-and-flip, and we use RandAugment\cite{cubuk2020randaugment} combined with the augmentation strategy in \cite{chen2020simple} as our another augmentation $DA_b$. More details about hyperparameters will be shown in Supplementary.

\begin{table}[]
\centering
\small
\renewcommand\arraystretch{1.1}
\renewcommand\tabcolsep{2.0pt}
\caption{Error rates for CIFAR-10 and on 5 different folds.}
\vspace{0.08in}
\begin{tabular}{c|ccc}
\toprule
                & \multicolumn{3}{c}{CIFAR-10}                      \\
                \midrule
Method          & 40 labels & 250 labels       & 4000 labels  \\
\midrule
$\pi$-model     & -         & $ 54.26 \pm 3.97$   &     $14.01 \pm 0.38$    \\
Pseduo-Labeling & -         & $49.78 \pm 0.38$   &    $16.09\pm 0.28$    \\
Mean Teacher    & -         &    $32.21 \pm 2.30$ &    $9.19 \pm 0.19$   \\
MixMatch        &    $47.54 \pm 11.50$   &    $11.05 \pm 0.86$   & $6.42 \pm 0.10$    \\
UDA             &    $29.05 \pm 5.93$  & $8.82 \pm 1.08$& $4.88 \pm 0.18$ \\
ReMixMatch      &   $19.10 \pm 9.64$ & $5.44 \pm 0.05$ & $4.72 \pm 0.13$ \\
FixMatch w.RA   & $13.81 \pm 3.37$ & \textbf{5.07 $\pm$ 0.65} & \textbf{4.26 $\pm$ 0.05}  \\
\midrule
SsCL     &  \textbf{10.29 $\pm$ 2.61}     &    \textbf{5.12 $\pm$ 0.41}      &   $4.51 \pm 0.13$   \\
\bottomrule
\end{tabular}
\label{tab:cifar}
\end{table}

\begin{table}[]
	\renewcommand\arraystretch{1.1}
	\renewcommand\tabcolsep{0.2pt}
	\caption{Semi-supervised learning results on ImageNet. We report top-1 and top-5 accuracy on the validation set with 1\% and 10\% labeled data. All models are based on ResNet50, and the results are all taken from their original papers except for MoCo v2 \cite{chen2020improved}, which is based on our re-implementation.}
	\vspace{0.08in}
	\small
	\begin{tabular}{c|c|c|c|c|c}
		\toprule
		&     & \multicolumn{2}{c}{1\% labels} & \multicolumn{2}{c}{10\% labels} \\
		Method & epochs      & top-1(\%)      & top-5(\%)      & top-1(\%)    & top-5(\%)         \\
		Supervised  & 20 & 25.4          & 48.4          & 56.4        & 80.4             \\
		\midrule
		\multicolumn{6}{c}{{Semi-supervised Based Method}}                              \\
		\midrule
		Pseudo-label\cite{lee2013pseudo}  & 100 & -             & 51.6          & -           & 82.4             \\
		VAT\cite{2019Virtual}   &-        & -             & 47            & -           & 83.4             \\
		UDA\cite{xie2019unsupervised}     &-      & -             & -             & 68.8        & 88.5             \\
		Fixmatch(w. RA)\cite{sohn2020fixmatch}   &400   & -             & -             & 71.5    & 89.1             \\
		S4L-Rotation\cite{zhai2019s4l} &200 & - & - & 53.4 & 83.8 \\
		\midrule
		\multicolumn{6}{c}{\emph{Self-supervised Based Method}}                              \\
		\midrule
		PIRL\cite{2019pretext}    &800      & 30.7          & 57.2          & 60.4        & 83.8             \\
		simCLR\cite{chen2020simple}    &1000    & 48.3          & 75.5          & 65.6        & 87.8             \\
		MoCo v2\cite{chen2020improved}   &800   & 52.4          & 78.4          & 65.3        & 86.6            \\
		BYOL\cite{grill2020bootstrap}    &1000      & 53.2          & 78.4          & 68.8        & 89.0             \\
		SwAV\cite{caron2020unsupervised}   &800       & 53.9          & 78.5          & 70.2        & 89.9             \\
		simCLR v2\cite{chen2020big}   &800  & 57.9          & 82.5          & 68.4        & 89.9             \\
		\midrule
		\multicolumn{6}{c}{\emph{Semi-supervised Contrastive Learning}}                              \\
		\midrule
		SsCL    &200      & 54.7  & 79.3   & 70.0        & 89.8    \\
		SsCL    &800   & \textbf{60.2} & \textbf{82.8} & \textbf{72.1}        & \textbf{90.9}    \\
		\bottomrule
	\end{tabular}
	\label{tab:imagenet}
\end{table}
\paragraph{Result} We report the result of different labels settings with our method in Table~\ref{tab:cifar}, and compare with some well-known approaches.
The improvement is more substantial when fewer labeled samples are available. We achieve $10.29\%$ error rate with only 40 labels, which is 3.5\% better than the previous best performed method Fixmatch\cite{sohn2020fixmatch} with more robust capability.

\subsection{ImageNet}
In this section, we evaluate our method on the large scale ImageNet dataset, which is challenging especially when extremely few shot labels are available. Following previous semi-supervised learning works on ImageNet \cite{chen2020simple,chen2020big, he2020momentum}, we conduct experiments with $1\%$ and $10\%$ labels for semi-supervised learning, using the same split of available labels for fair comparisons. Besides, different from the previous contrastive learning based methods that first conduct pretraining and then followed by few shot fine tuing, our method is an end-to-end framework.
\paragraph{Implementation Details}
For model training, we use ResNet-50\cite{He2016Deep} as the backbone, and use the SGD optimizer with weight decay of 0.0001 and momentum as 0.9. The model is trained on 32 Tesla V100 GPUs with batch size of 1024 for unlabeled data and 256 for labeled data, and the initial learning rate is 0.4 with cosine decay schedule. The data augmentation follows the same strategy as \cite{chen2020simple}. More hyperparameters details are shown in Supplementary.
\paragraph{Result} We compare our method with several well-known semi-supervised learning approaches, which can be roughly categorized into two aspects,, \emph{i.e.,} the conventional semi-supervised learning strategy that makes use of cross entropy loss for data mining, as well as contrastive learning based methods that first pre-train the models without using any image-level labels, followed by directly fine-tuning with few shot labeled data. As shown in Table~\ref{tab:imagenet}, We achieve top-1 accuracies of $60.2\%$ and $72.1\%$ with $1\%$ and $10\%$ labeled data, respectively, which significantly improves the baseline MoCo v2 \cite{chen2020improved}. Under $1\%$ setting, our method is also better than previous best performed method simCLR v2. It should be noted that simCLR v2 \cite{chen2020big} makes use of extra tricks such as more MLP layers, which has been demonstrated to be effective for semi-supervised learning, while we simply follow the MoCo v2 baseline that makes use of two MLP layers. Under $10\%$ setting, our method is $0.6\%$ better than the previous best performed Fixmatch in top-1 accuracy, and $1.8\%$ higher in top-5 accuracy. We also find that previous semi-supervised works rarely report accuracy with $1\%$ labels due to their poor performance. The reason is that the extremely few labeled data is too biased to represent the corresponding class, and such error can be magnified during the data mining procedure using cross entropy loss, while our method is robust to the bias, and especially effective when few shot labels are available.

\begin{table}[]
	\renewcommand\arraystretch{1.2}
	\setlength{\tabcolsep}{1.6mm}
	\caption{Transfer learning accuracy (\%) on COCO detection (averaged over 5 trials).}
	\vspace{0.08in}
	\centering
	\begin{tabular}{l|cccccc}
		\hline
		\multirow{3}{*}{Method} & \multicolumn{6}{c}{Mask R-CNN, R50-FPN, Detection}  \\ \cline{2-7}
		& \multicolumn{6}{c}{$1\times$ schedule}  \\ \cline{2-7}
		& AP$^{bb}$    & AP$^{bb}_{50}$ & AP$^{bb}_{75}$ & AP$_S$  & AP$_M$ &AP$_L$  \\ \hline
		Supervised              & 38.9  & 59.6 & 42.0 & 23.0     &42.9  &49.9      \\ \hline
		MoCo v2  \cite{chen2020improved}               & 39.2  & 59.9 & 42.7 & 23.8     &  42.7 &50.0  \\ \hline
		Ours 10\%   & \textbf{40.0}  & \textbf{61.6} & \textbf{43.4} & \textbf{23.9} &\textbf{43.9} &\textbf{51.7}\\ \hline
	\end{tabular}
	\label{tab: coco_detec}
\end{table}

\begin{table}[]
	\centering
	\caption{The intra-class similarity of each method on validation set after training 400 epochs.}
	\vspace{0.08in}
	\small
	\renewcommand\arraystretch{1.1}
	\setlength{\tabcolsep}{1.6mm}
	\begin{tabular}{c|c|c|c|c}
		\toprule
		& SwAV & Deepcluster v2 & Ours 1\%& Ours 10\% \\
		\midrule
		intra-class &0.46 &0.43 & \textbf{0.54} & \textbf{0.63}\\
		\bottomrule
	\end{tabular}   
	\label{tab:similarity}
\end{table}

\begin{table}[]
	\centering
	\caption{The influence of similarity alignment, w. and w/o which means with and without this operation respectively.}
	\vspace{0.08in}
	\small
	\renewcommand\arraystretch{1.2}
	\setlength{\tabcolsep}{1.2mm}
	\begin{tabular}{c|c|c|c|c}
		\toprule
		Pseudo label acc. & Nearest class acc. & Overlap & w. & w/o.  \\
		\midrule
		53.2\% & 45.3\% & 60.1\%  & 50.8\% & 48.4\%  \\
		\bottomrule
	\end{tabular}
	\label{tab:FA}
\end{table}

\subsection{Downstream Tasks}

We further evaluate our learned representation on downstream object detection task to uncover the transferability. Following \cite{he2020momentum}, we choose Mask R-CNN\cite{2017Mask} with FPN\cite{lin2017feature} as the backbone, and fine-tune all the parameters on the training set, then evaluate it on the validation set of COCO 2017. We report our performance under 1$\times$ schedules in Table~\ref{tab: coco_detec}. It can be seen that our method consistently outperforms the supervised pretrained model and MoCo v2.

\subsection{Study and Analysis}
\begin{figure*}
    \centering
    \includegraphics[width=\textwidth]{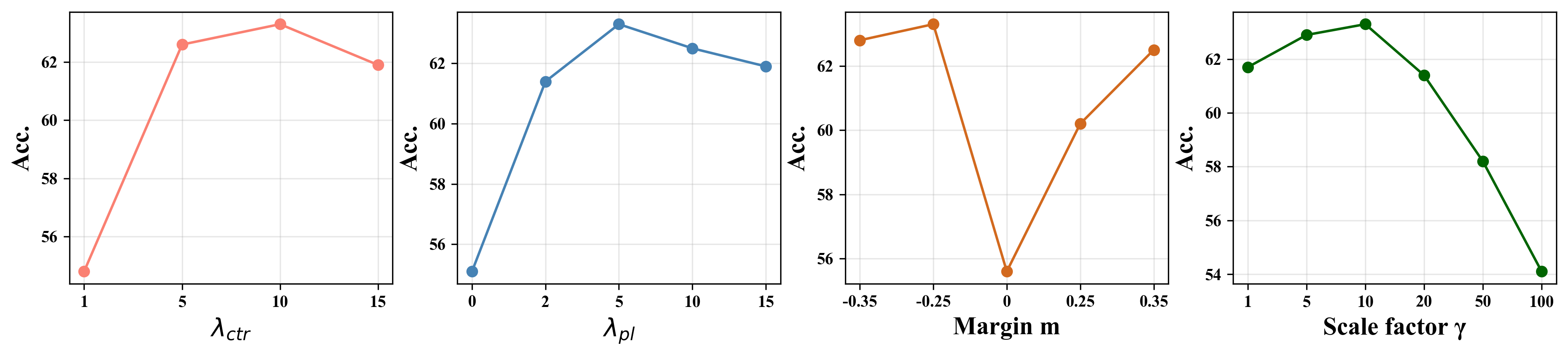}
    \caption{The influences of different parameters. From left to right (1) Varing the weight $\lambda_{ctr}$ for contrastive loss. (2) Varing the weight $\lambda_{pl}$ for pseudo label loss. (3) Varing margin $m$ for contrastive loss. (4) Varing $\gamma$ which controls the scale of contrastive loss.}
    \label{F:hyparameters}
\end{figure*}

We give an in-depth analysis of our design for semi-supervised contrastive learning, and demonstrate the benefits of them through comparative experiments. Unless specified, the results are based on ImageNet with 1\% labels training for 100 epochs.
\paragraph{Distinguish with clustering method} Our approach targets at pulling nearest class samples together with calculated prototypes, this is similar to the idea of some clustering based methods. So we use intra-class similarity as an indicator to measure our learned features. We compare our method with two classic clustering-based learning methods, SwAV\cite{caron2020unsupervised} and Deepcluster v2 \cite{caron2018deep}, and the results are shown in Table~\ref{tab:similarity}. The intra-class similarity is calculated by averaging cosine distance among all intra-class pairwise samples, and we report the average of 1000 classes on the ImageNet validation set. Our method consistently outperforms these two methods both in 1\% and 10\% labels available, which validates that it is better for discrimination when more class priors are available.
\paragraph{Co-calibration helps for better representation} Table~\ref{tab:FA} provides some interesting results. If we use pretrained backbone and $fc$ head to predict unlabeled training data, we can get top-1 accuracy of 53.2\%, while the accuracy drops to 45.3\% if we use the nearest class in feature space as its label. However, the overlap of their correct prediction is only 60.1\%, in another word, most of the labels that one method predicts wrong are correct in another method, which is the motivation of our co-calibration mechanism. We believe this will help the model learn complementary information during training, and it can be seen that our method gains 2.4\% benefits with this operation.

\begin{figure}
    \centering
    \includegraphics[width=\columnwidth]{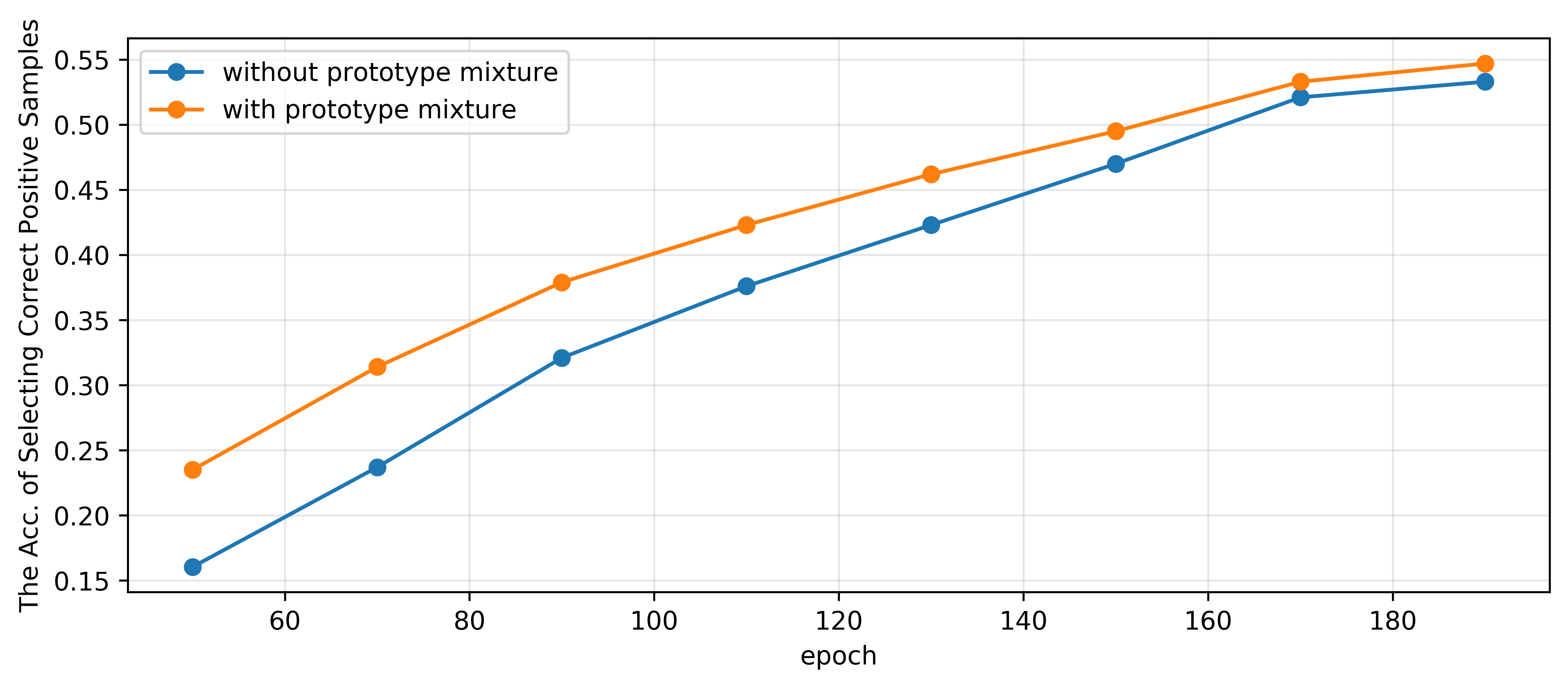}
    \caption{The top-1 accuracy comparisons with and without using prototype mixture. }
    \label{fig:proto}
\end{figure}
\paragraph{Prototype mixture helps for better prediction}
We calculate the average accuracy of the training samples which can get the correct positive samples by assigning the nearest category before and after adding the mixture module. The results are shown in Fig.~\ref{fig:proto}. We can see that this module brings about 8\% accuracy improvement at most. In addition, the model with prototype mixture always keeps high accuracy in selecting positive samples, and more correct positive samples are undoubtedly beneficial to improve the performance.



\section{Ablation study}
In this section, we conduct detailed ablation studies to inspect how each hyperparamater affects the performance. For efficiency, unless specified, all experiments are conducted with 100 epochs using a randomly selected subset of 100 categories in ImageNet with 1\% labels. We report top-1 accuracy in each experiment, where we achieve 63.3\% with our default settings. Besides, We also re-implement MoCo v2\cite{chen2020improved}, using it as our $\mathcal{L}_{ctr}$ under this setting as our baseline for fair comparisons. 

\subsection{Hyper-parameters of Loss Function}\label{sec:hyperparam}
We first analyze the impact of several hyper-parameters, which includes pseudo label loss weight $\lambda_{pl}$, contrastive loss weight $\lambda_{ctr}$, scale factor $\gamma$, weight factor $\alpha_p$ and the margin value $m$.

\paragraph{The scale factor $\gamma$} The effect of this coefficient is equal to $\tau$ in \cite{chen2020improved,chen2020simple}. With too large $\gamma$, it will penalize more about the discrepancy, which will break the semantic structure of the embedding distribution, as it is shown in Figure~\ref{F:hyparameters}. We set it in 10 in our ImageNet experiment. 
\paragraph{Loss weight} We observe the influence of two weights $\lambda_{pl}$ and $\lambda_{ctr}$ on the performance and report the result in the Figure~\ref{F:hyparameters}, where $\lambda_{pl}=5$ and $\lambda_{ctr}=10$ gives the best performance. But in the 10\% labels setting, we need to increase the $\lambda_{pl}$ to 10 and decrease the $\lambda_{ctr}$ to 5 to get the best result. We think it is because the fewer labeled data, the worse the performance on predicting pseudo labels, which need a larger $\lambda_{ctr}$ to strengthen the contrastive regularization.
\paragraph{The weight factor $\alpha_p$} The comparison results of introducing $\alpha_p$ are shown in Table~\ref{tab:alpha}. It can be seen that introducing adaptive weight $\alpha_p$ is beneficial for improving classification accuracy. It brings about $4.9\%$  gains with fixed $\alpha_p$ (set as 1). This is mainly due to the reason that $\alpha_p$ assigns a lower penalty to the positive samples that are less similar to the assigned class, which reduces the influence of these relatively noisy samples during model training.
\paragraph{The margin value $m$} Inspired by \cite{xie2020delving}, we set $m$ as negative because the pseudo labels are not always correct during training. Figure~\ref{F:hyparameters} shows the result of different settings, we observe that both positive and negative margins can improve the quality of learned features, 
so it is beneficial to design a less stringent decision boundary for better class separability.

\begin{table}[]
\centering
\caption{Impact of different settings of $\alpha_p$.}
\vspace{0.08in}
\begin{tabular}{c|c|c}
\toprule
\multirow{2}{*}{Method} & \multicolumn{2}{c}{Accuracy(\%)} \\
                        & top-1            & top-5           \\
\midrule
w/o $\alpha_p$                       & 58.4           & 81.2          \\
w/  $\alpha_p$                       & 63.3            & 85.7           \\
\bottomrule
\end{tabular}
\label{tab:alpha}
\end{table}

\begin{table}[]
	\centering
	\small
	\caption{The comparison results of adding different number of positive samples for contrastive learning.}
	\vspace{0.08in}
	\renewcommand\arraystretch{1.2}
	\setlength{\tabcolsep}{2.4mm}
	\begin{tabular}{c|c|c|c|c}
		\toprule
		Number & 0 & 2 & 3 & 4  \\
		\midrule
		Top-1 Acc.(\%) & 56.9 & 62.7  & 63.3 & 62.8  \\
		\bottomrule
		
	\end{tabular}
	\label{tab:positive-sample}
\end{table}
\vspace{0.5cm}

\subsection{The Number of Positive Samples}\label{sec:pos_num}

Table~\ref{tab:positive-sample} shows the results of adding the different number of positive samples. 0 here means the original contrastive loss\cite{chen2020improved} that does not pull semantically similar images. It can be seen that adding extra positive samples benefits class separability, which significantly boosts the performance. 

\section{Conclusion}
This paper proposed a semi-supervised contrastive learning framework, which incorporates contrastive loss into semi-supervised learning, and jointly optimizes the network in an end-to-end way. The main contribution is that we introduce a co-calibration mechanism to interchange predictions between the two branches, which we find is complementary and thus avoid trapped in local minimum. Experiments conducted on several datasets demonstrate the superior performance of the proposed method, especially when extremely few labeled samples are available. 

\appendix
\section{Appendix}
\subsection{More Detail in Training Process}
\paragraph{Hyperpatrameters Detail.} Here we provide a complete list of the hyperparameters we use in our experiment in Table~\ref{t:parameters}. Note that we did ablation studies for some of these parameters in previous sections, including loss weight $\lambda_{ctr}$ and $\lambda_{cls}$, scale factor $\gamma$ and margin $m$, As to other parameters we just follow the settings in previous works\cite{sohn2020fixmatch,he2020momentum}. 

\paragraph{Derivation Process of Loss Function.} Here we provide the specific derivation process of our loss function corresponding to  Eq.~(9) in the text. The details are shown in the following: 

 \begin{small}
\begin{align}
    \mathcal{L}_{ctr} &= \log (1 + \sum\limits_{{k^ - }} {\exp (\gamma({s_n} + m))} \sum\limits_{{k^ + }}\exp ( - \gamma \alpha_p {s_p})) \notag \\
    &\approx max(\log \sum\limits_{{k^ - }} {\exp (\gamma({s_n} + m))} \sum\limits_{{k^ + }}\exp ( - \gamma \alpha_p{s_p}),0) \notag \\
    &= max(\log \sum\limits_{{k^ - }} {\exp (\gamma({s_n} + m))} \log \sum\limits_{{k^ + }}\exp ( - \gamma \alpha_p {s_p}),0) \notag \\
    &= max(\gamma(L_{LSE}(s_n)+m-L_{-LSE}(\alpha_p s_p)),0) \notag \\
    &\approx max(\gamma(max(s_n)-min(\alpha_p s_p)+m),0)
    \label{eq:min_max}
\end{align}
\end{small}

\begin{figure*}
\centering
\includegraphics[width=\textwidth]{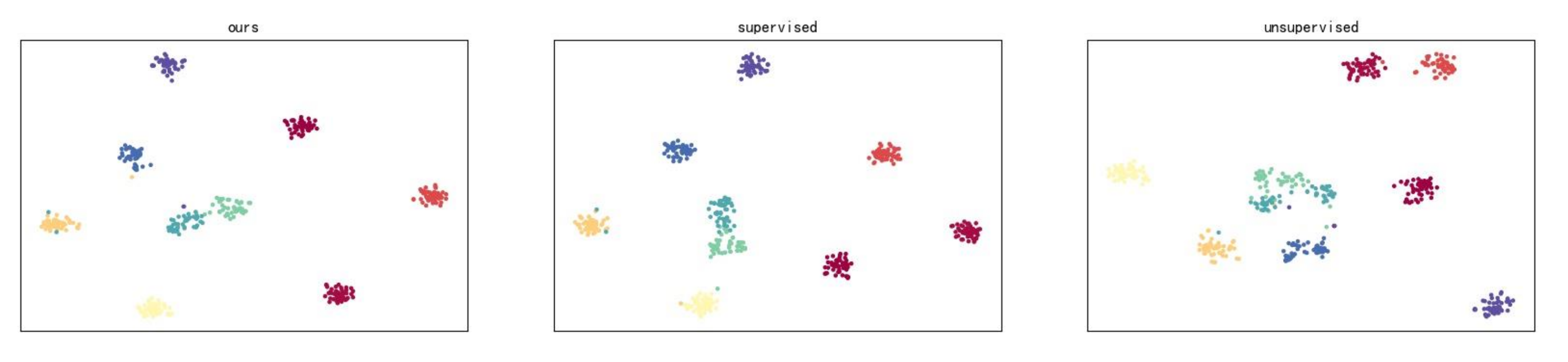}
\caption{$t-sne$ visualization of representation learned by SsCL, supervised and unsupervised.}
\vspace{-0.4cm}
\label{fig:t-sne}
\end{figure*}

\section{Transfer Learning}
We evaluated the quality of our representation for transfer learning in two settings: linear evaluation, where a logistic regression classifier is trained to classify a new dataset based on the representation learned by ScCL
on ImageNet, and fine-tuning, where we allow all weights to vary during training. 
In both cases, we follow the settings of \cite{chen2020simple,2019Revisiting}.
\subsection{Dataset}
We use four tiny scale dataset in this experiment, including CIFAR-10/100\cite{AlexLearning2009},  FGVC Aircraft\cite{maji2013fine} and Standford Cars\cite{krause2013collecting}. We report top-1 accuracy for these datasets with our 10\% labels 800 epochs pre-trained model. 
\subsection{Implemental Detail}
For linear evaluation, we simply follow the settings used in moco v2\cite{chen2020improved}, where learning rate is set in 30 and training for 100 epochs. While in the fine-tuning task, we use the SGD optimizer to fine-tune all network parameters.  The learning rate is set as $1e-3$ for backbone and $1$ for the classifier, and the total epoch is 120 in the second stage and the learning rate is decayed by 0.1 in 50,70 and 90 epoch.  

\begin{table}[]
	\caption{Hyperparameters for SsCL in our experiments.}
	\renewcommand\arraystretch{1.2}
	\renewcommand\tabcolsep{3.3pt}
	\vspace{0.08in}
	\begin{tabular}{cccccccccccc}
		\toprule
		& $\lambda_{ctr}$ &$\lambda_{cls}$ & B & $\mu$  & K & $\tau$ & $\gamma$ & $m$  \\
		\midrule
		CIFAR-10      &  1  & 1    &    64         &      4       & 4096  &   0.95  &   5  & -0.25   \\
		ImageNet-1\%  & 10  & 5    &       \multirow{2}{*}{64}      &    \multirow{2}{*}{4}         & \multirow{2}{*}{65536}   &  \multirow{2}{*}{0.6}   &     \multirow{2}{*}{10} & \multirow{2}{*}{-0.25} \\
		ImageNet-10\% & 5  & 10    &             &             &   &     &   		\\
		\bottomrule
	\end{tabular}
	\label{t:parameters}
\end{table}
\subsection{Baseline}
We compare our approach with the supervised and unsupervised method. The supervised method is trained on ImageNet with standard cross-entropy loss, while we use well-known methods simCLR\cite{chen2020simple} and MoCo v2\cite{chen2020improved} as our unsupervised baseline. Their result are all taken from \cite{chen2020simple} except for MoCo v2, which is based on our re-implementation. Besides, in the fine-tuning task, we add a setting by training a model from random initialization for comparison.

\subsection{Result}

It can be seen from Table~\ref{t:transfer} that the  supervised method has a clear advantage over self-supervised and semi-supervised training in linear evaluation. By introducing more label prior information, which is beneficial for the separation of representation, we outperform the self-supervised based method on almost all datasets with both linear evaluation and fine-tuning task. Besides, we achieve superior performance to the supervised method in almost all datasets with fine-tuning, which also demonstrates the transfer ability of our model.

\section{Visualization of Feature Representation}
We visualize the embedding feature to better understand the advantage of our semi-supervised contrastive training method. We randomly choose ten classes from the
validation set and provide the $t-sne$ visualization of feature representation generated by ScCL, supervised training and MoCo v2\cite{he2020momentum}. The results are shown in Figure~\ref{fig:t-sne}, the same color denotes features with the same label.  It can be seen that ScCL takes on higher aggregation property compared with MoCo v2, and is slightly worse than the supervised method because it makes use of all labels.

{\small
\bibliographystyle{ieee_fullname}
\bibliography{egbib}

\begin{thebibliography}{10}\itemsep=-1pt

\bibitem{arazo2020pseudo}
Eric Arazo, Diego Ortego, Paul Albert, Noel~E O’Connor, and Kevin McGuinness.
\newblock Pseudo-labeling and confirmation bias in deep semi-supervised
  learning.
\newblock In {\em 2020 International Joint Conference on Neural Networks
  (IJCNN)}, pages 1--8. IEEE, 2020.

\bibitem{berthelot2019remixmatch}
David Berthelot, Nicholas Carlini, Ekin~D Cubuk, Alex Kurakin, Kihyuk Sohn, Han
  Zhang, and Colin Raffel.
\newblock Remixmatch: Semi-supervised learning with distribution alignment and
  augmentation anchoring.
\newblock {\em arXiv preprint arXiv:1911.09785}, 2019.

\bibitem{caron2018deep}
Mathilde Caron, Piotr Bojanowski, Armand Joulin, and Matthijs Douze.
\newblock Deep clustering for unsupervised learning of visual featrues.
\newblock 2018.

\bibitem{caron2020unsupervised}
Mathilde Caron, Ishan Misra, Julien Mairal, Priya Goyal, Piotr Bojanowski, and
  Armand Joulin.
\newblock Unsupervised learning of visual features by contrasting cluster
  assignments.
\newblock 33, 2020.

\bibitem{chen2020simple}
Ting Chen, Simon Kornblith, Mohammad Norouzi, and Geoffrey Hinton.
\newblock A simple framework for contrastive learning of visual
  representations.
\newblock {\em arXiv preprint arXiv:2002.05709}, 2020.

\bibitem{chen2020big}
Ting Chen, Simon Kornblith, Kevin Swersky, Mohammad Norouzi, and Geoffrey
  Hinton.
\newblock Big self-supervised models are strong semi-supervised learners.
\newblock {\em arXiv preprint arXiv:2006.10029}, 2020.

\bibitem{chen2020improved}
Xinlei Chen, Haoqi Fan, Ross Girshick, and Kaiming He.
\newblock Improved baselines with momentum contrastive learning.
\newblock {\em arXiv preprint arXiv:2003.04297}, 2020.

\bibitem{cubuk2020randaugment}
Ekin~D Cubuk, Barret Zoph, Jonathon Shlens, and Quoc~V Le.
\newblock Randaugment: Practical automated data augmentation with a reduced
  search space.
\newblock In {\em Proceedings of the IEEE/CVF Conference on Computer Vision and
  Pattern Recognition Workshops}, pages 702--703, 2020.

\bibitem{dosovitskiy2015discriminative}
Alexey Dosovitskiy, Philipp Fischer, Jost~Tobias Springenberg, Martin
  Riedmiller, and Thomas Brox.
\newblock Discriminative unsupervised feature learning with exemplar
  convolutional neural networks.
\newblock 38(9):1734--1747, 2015.

\bibitem{grill2020bootstrap}
Jean-Bastien Grill, Florian Strub, Florent Altch{\'e}, Corentin Tallec, Pierre
  Richemond, Elena Buchatskaya, Carl Doersch, Bernardo Avila~Pires, Zhaohan
  Guo, Mohammad Gheshlaghi~Azar, et~al.
\newblock Bootstrap your own latent-a new approach to self-supervised learning.
\newblock 33, 2020.

\bibitem{guillaumin2010multimodal}
Matthieu Guillaumin, Jakob Verbeek, and Cordelia Schmid.
\newblock Multimodal semi-supervised learning for image classification.
\newblock In {\em 2010 IEEE Computer society conference on computer vision and
  pattern recognition}, pages 902--909. IEEE, 2010.

\bibitem{he2020momentum}
Kaiming He, Haoqi Fan, Yuxin Wu, Saining Xie, and Ross Girshick.
\newblock Momentum contrast for unsupervised visual representation learning.
\newblock In {\em Proceedings of the IEEE/CVF Conference on Computer Vision and
  Pattern Recognition}, pages 9729--9738, 2020.

\bibitem{2017Mask}
Kaiming He, Georgia Gkioxari, Piotr Dollár, and Ross Girshick.
\newblock Mask r-cnn.
\newblock {\em IEEE Transactions on Pattern Analysis and Machine Intelligence},
  2017.

\bibitem{He2016Deep}
Kaiming He, Xiangyu Zhang, Shaoqing Ren, and Jian Sun.
\newblock Deep residual learning for image recognition.
\newblock 2016.

\bibitem{hjelm2018learning}
R~Devon Hjelm, Alex Fedorov, Samuel Lavoie-Marchildon, Karan Grewal, Phil
  Bachman, Adam Trischler, and Yoshua Bengio.
\newblock Learning deep representations by mutual information estimation and
  maximization.
\newblock {\em arXiv preprint arXiv:1808.06670}, 2018.

\bibitem{2019pretext}
Misra Ishan and van~der Maaten~Laurens.
\newblock Self-supervised learning of pretext-invariant representations.
\newblock 2020.

\bibitem{long2020self}
Longlong Jing and Yingli Tian.
\newblock Self-supervised visual featrue learning with deep neural networks: A
  survey.
\newblock 2020.

\bibitem{kingma2014semi}
Durk~P Kingma, Shakir Mohamed, Danilo Jimenez~Rezende, and Max Welling.
\newblock Semi-supervised learning with deep generative models.
\newblock 27:3581--3589, 2014.

\bibitem{lee2013pseudo}
Dong-Hyun Lee.
\newblock Pseudo-label: The simple and efficient semi-supervised learning
  method for deep neural networks.
\newblock In {\em Workshop on challenges in representation learning, ICML},
  volume~3, 2013.

\bibitem{lin2017feature}
Tsung-Yi Lin, Piotr Doll{\'a}r, Ross Girshick, Kaiming He, Bharath Hariharan,
  and Serge Belongie.
\newblock Feature pyramid networks for object detection.
\newblock In {\em Proceedings of the IEEE conference on computer vision and
  pattern recognition}, pages 2117--2125, 2017.

\bibitem{2019Virtual}
Takeru Miyato, Shin~Ichi Maeda, Masanori Koyama, and Shin Ishii.
\newblock Virtual adversarial training: A regularization method for supervised
  and semi-supervised learning.
\newblock 41(8):1979--1993, 2019.

\bibitem{oliver2018realistic}
Avital Oliver, Augustus Odena, Colin~A Raffel, Ekin~Dogus Cubuk, and Ian
  Goodfellow.
\newblock Realistic evaluation of deep semi-supervised learning algorithms.
\newblock pages 3235--3246, 2018.

\bibitem{oord2018representation}
Aaron van~den Oord, Yazhe Li, and Oriol Vinyals.
\newblock Representation learning with contrastive predictive coding.
\newblock {\em arXiv preprint arXiv:1807.03748}, 2018.

\bibitem{rasmus2015semi}
Antti Rasmus, Mathias Berglund, Mikko Honkala, Harri Valpola, and Tapani Raiko.
\newblock Semi-supervised learning with ladder networks.
\newblock pages 3546--3554, 2015.

\bibitem{sajjadi2016regularization}
Mehdi Sajjadi, Mehran Javanmardi, and Tolga Tasdizen.
\newblock Regularization with stochastic transformations and perturbations for
  deep semi-supervised learning.
\newblock pages 1163--1171, 2016.

\bibitem{sohn2020fixmatch}
Kihyuk Sohn, David Berthelot, Chun-Liang Li, Zizhao Zhang, Nicholas Carlini,
  Ekin~D Cubuk, Alex Kurakin, Han Zhang, and Colin Raffel.
\newblock Fixmatch: Simplifying semi-supervised learning with consistency and
  confidence.
\newblock {\em arXiv preprint arXiv:2001.07685}, 2020.

\bibitem{sukhbaatar2014training}
Sainbayar Sukhbaatar, Joan Bruna, Manohar Paluri, Lubomir Bourdev, and Rob
  Fergus.
\newblock Training convolutional networks with noisy labels.
\newblock {\em arXiv preprint arXiv:1406.2080}, 2014.

\bibitem{sun2020circle}
Yifan Sun, Changmao Cheng, Yuhan Zhang, Chi Zhang, Liang Zheng, Zhongdao Wang,
  and Yichen Wei.
\newblock Circle loss: A unified perspective of pair similarity optimization.
\newblock In {\em CVPR}, pages 6398--6407, 2020.

\bibitem{tarvainen2017mean}
Antti Tarvainen and Harri Valpola.
\newblock Mean teachers are better role models: Weight-averaged consistency
  targets improve semi-supervised deep learning results.
\newblock pages 1195--1204, 2017.

\bibitem{tian2019contrastive}
Yonglong Tian, Dilip Krishnan, and Phillip Isola.
\newblock Contrastive multiview coding.
\newblock {\em arXiv preprint arXiv:1906.05849}, 2019.

\bibitem{wu2018unsupervised}
Zhirong Wu, Yuanjun Xiong, Stella~X Yu, and Dahua Lin.
\newblock Unsupervised feature learning via non-parametric instance
  discrimination.
\newblock pages 3733--3742, 2018.

\bibitem{xie2020delving}
Jiahao Xie, Xiaohang Zhan, Ziwei Liu, Yew~Soon Ong, and Chen~Change Loy.
\newblock Delving into inter-image invariance for unsupervised visual
  representations.
\newblock {\em arXiv preprint arXiv:2008.11702}, 2020.

\bibitem{xie2019unsupervised}
Qizhe Xie, Zihang Dai, Eduard Hovy, Minh-Thang Luong, and Quoc~V Le.
\newblock Unsupervised data augmentation for consistency training.
\newblock {\em arXiv preprint arXiv:1904.12848}, 2019.

\bibitem{zagoruyko2016wide}
Sergey Zagoruyko and Nikos Komodakis.
\newblock Wide residual networks.
\newblock In {\em British Machine Vision Conference 2016}. British Machine
  Vision Association, 2016.

\bibitem{zhai2019s4l}
Xiaohua Zhai, Avital Oliver, Alexander Kolesnikov, and Lucas Beyer.
\newblock S4l: Self-supervised semi-supervised learning.
\newblock pages 1476--1485, 2019.

\bibitem{zhang2018mixup}
Hongyi Zhang, Moustapha Cisse, Yann~N Dauphin, and David Lopez-Paz.
\newblock mixup: Beyond empirical risk minimization.
\newblock 2018.

\bibitem{zhang2018generalized}
Zhilu Zhang and Mert Sabuncu.
\newblock Generalized cross entropy loss for training deep neural networks with
  noisy labels.
\newblock pages 8778--8788, 2018.

\end{thebibliography}
}

\end{document}